%% file: egpaper.tex
\documentclass[10pt,twocolumn,letterpaper]{article}

\usepackage{wacv}
\usepackage{times}
\usepackage{epsfig}
\usepackage{graphicx}
\usepackage{amsmath}
\usepackage{amssymb}
\usepackage{booktabs}
\usepackage{xcolor}         
\usepackage{multirow}
\usepackage{subcaption}
\usepackage{comment}
\usepackage{algorithm}
\usepackage[noend]{algpseudocode}

\wacvfinalcopy 

\ifwacvfinal
\usepackage[breaklinks=true,bookmarks=false]{hyperref}
\else
\usepackage[pagebackref=true,breaklinks=true,colorlinks,bookmarks=false]{hyperref}
\fi

\input{macros}
\newcommand{\shortname}{NeuralODF\xspace}
\title{\shortname: Learning Omnidirectional Distance Fields for 3D Shape Representation}

\author{
Trevor Houchens\thanks{Authors contributed equally to this work.} \\
Brown University\\
{\tt\small trevor\_houchens@brown.edu}
\and
Cheng-You Lu$^{\ast}$ \\
Brown University\\
{\tt\small cheng-you\_lu@brown.edu} 
\and
Shivam Duggal \\
Carnegie Mellon University \\
{\tt\small sduggal@andrew.cmu.edu}
\and
Rao Fu \\
Brown University\\
{\tt\small rao\_fu@brown.edu}
\and
Srinath Sridhar \\
Brown University \\
{\tt\small srinath@brown.edu}
}

\begin{document}

\maketitle
\thispagestyle{empty}

\input{content/00_abstract}
\input{content/01_intro}

\input{content/02_relwork}
\input{content/03_method}

\input{content/04_experiments}

\input{content/05_endmatter}
{\small
\bibliographystyle{ieee_fullname}
\bibliography{egbib}
}

\clearpage
\input{content/99_appendix}

\end{document}

%% file: macros.tex
\usepackage{cleveref}

\usepackage{wrapfig}
\usepackage{ifthen}

\usepackage{xspace}

\ifx \cvprsubmission \undefined
    
\else
	
\fi

\newcommand{\parahead}[1]{\noindent\textbf{#1:}\ }

\newenvironment{packed_itemize}
{\begin{itemize}
    \setlength{\itemsep}{1pt}
    \setlength{\parskip}{0pt}
    \setlength{\parsep}{0pt}
}{\end{itemize}}

\newcommand{\filluptopage}[1]{%
  \clearpage
  \loop\ifnum\value{page}<#1\relax
    \null\clearpage
  \repeat
  \loop\ifnum\value{page}=#1\relax
    \null\clearpage
  \repeat
}

\makeatletter
\def\blfootnote{\xdef\@thefnmark{}\@footnotetext}
\makeatother

%% file: content/00_abstract.tex
\begin{abstract}
%

The 3D shape of objects is typically represented as meshes, point clouds, voxel grids, level sets, or depth images.
Each representation is suited for different tasks thus making the transformation of one representation into another (forward map) an important and common problem.
We propose Omnidirectional Distance Fields (ODFs), a new 3D shape representation that encodes geometry by storing the distance to the object's surface from any 3D position in any viewing direction.
Since \emph{rays} are the fundamental unit of an ODF, it enables easy transformation to and from common 3D representations like meshes or point clouds.
Different from level set methods that are limited to representing closed surfaces, ODFs are unsigned and can thus model open surfaces (\eg~garments).
We demonstrate that ODFs can be effectively learned with a neural network (\shortname) despite the inherent discontinuities at occlusion boundaries.
We also introduce efficient forward mapping algorithms for transforming ODFs to and from common 3D representations.
Experiments demonstrate that \shortname can learn to capture high-quality shape by overfitting to a single object, and also learn to generalize on common shape categories.
\end{abstract}

%% file: content/01_intro.tex
\section{Introduction}
\label{sec:intro}
Representing the 3D shape of objects and scenes is a fundamental problem in visual computing.
Shape representations such as point clouds, meshes, voxel grids, level sets, depth images and multi-view images are widely used in 3D scene reconstruction, rendering, design, or 3D printing.
Each representation has its own advantages and issues for use in downstream tasks, so transforming one to another is an important
task.

Recently, a class of coordinate-based neural networks, known as \emph{neural fields}~\cite{xie2021neural} or \emph{neural implicit functions}, have been used to learn these 3D shape representations~\cite{park2019deepsdf,mescheder2019occupancy,groueix2018papier,yang2018foldingnet,chen2019learning}.
These neural \emph{shape} fields parametrize 3D shape using coordinates in a bounded volume or a surface, and have been shown to be useful for shape modeling, and reconstructing shapes from images or other partial inputs.
Existing neural shape fields are typically trained by sampling the underlying 3D representation they model (\eg~SDFs).
To produce suitable outputs for downstream tasks (\eg~meshes), \textbf{forward maps}~\cite{xie2021neural} are used.
Common examples of forward maps include marching cubes~\cite{lorensen1987marching} for extracting meshes from level sets, ray marching~\cite{perlin1989hypertexture} for rendering (depth) images from level sets, and surface splatting~\cite{zwicker2001surface} for rendering from point clouds.
However, forward maps to transform neural shape fields to common representations can be complex and computationally expensive.
Moreover, many extant neural shape fields~\cite{park2019deepsdf,mescheder2019occupancy} can only model \emph{watertight (closed) surfaces}---shapes with a clearly defined inside and outside---preventing their use in modeling open surface shapes like garments.

\begin{figure*}[t]
  \centering
  \includegraphics[width=0.85\textwidth]{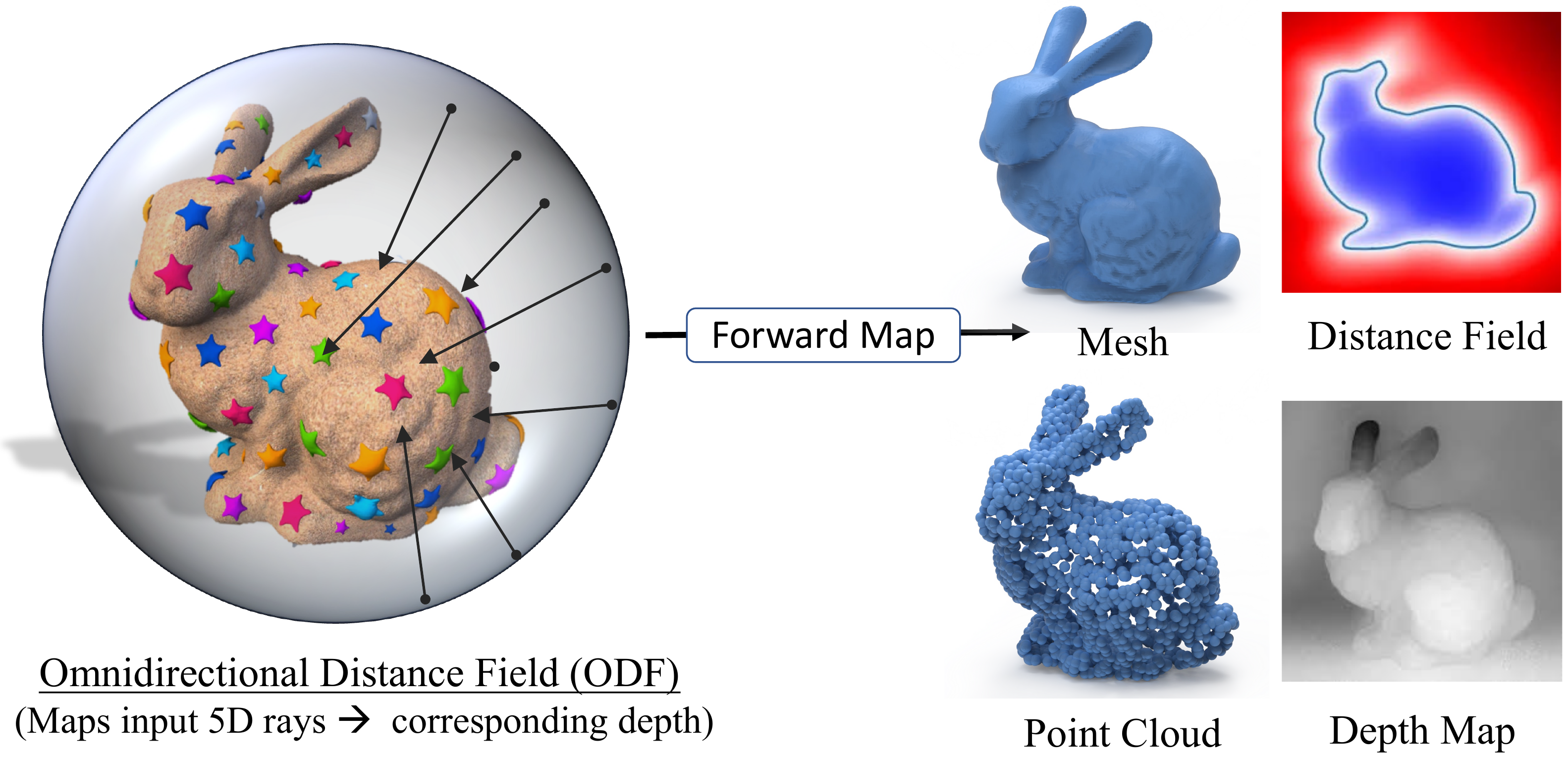}
  \caption{We introduce \textbf{Omnidirectional Distance Field (ODF)}, a new 3D shape representation that takes a 5D input, $\vec r = (x, \hat{w})$, specifying a ray originating at a 3D location $x$ and oriented along a direction $\hat{w}$.
For each such ray, an ODF outputs the distance to the surface of the 3D shape along the input ray.
  We show that \textbf{\shortname} can implicitly learn an ODF when trained on rays sampled within a region around the shape (sphere shown above).
  We also show that \shortname can be easily transformed to common shape representations like meshes, points clouds, or depth maps using computationally-efficient forward maps~\cite{xie2021neural}.
  }
  \vspace{-0.2in}
  \label{fig:teaser}
\end{figure*}

In this paper, we address these limitations by proposing \textbf{Omnidirectional Distance Field (ODF)}, a new 3D shape representation that allows more efficient forward mapping to common shape representations while capturing both open and watertight surfaces.
An ODF implicitly encodes object geometry by storing the distance to the object's surface from any 3D position, $x$, and in any viewing direction, $\hat{w}$.
For any open or closed surface $\partial \Omega$ we formally define an ODF as
\begin{equation*}
 d(x, \hat{w})=\min(\lVert x-x_I\rVert_2),\text{ }x_I\in\partial\Omega,
\end{equation*}
where $x_I=x+t\cdot\hat{w}$ with $t\geq0$.
\textbf{The fundamental building block of an ODF is a ray}.
Intuitively, an ODF stores the distance along the rays cast towards a surface in any direction (omnidirectional) and originating at any 3D point (see \Cref{fig:teaser}).
ODFs generalize unsigned distance fields \cite{chibane2020neural} to all directions, and can be used to represent open surfaces.
However, an ODF is parametrized in 5D (3D position and 2D direction) making it inefficient for storage and high-resolution querying.
Therefore, we show that a neural network, \textbf{\shortname}, can implicitly learn an ODF for efficient storage and fast querying.
At inference, rather than estimate the ray distnaces in a single forward pass, we use a recursive approach to estimate higher-quality shapes.
Since ODFs are inherently discontinuous at occlusion boundaries, we estimate visibility enabling us to handle discontinuities.
%

The omnidirectional distance information in a \shortname enables us to build forward maps to extract meshes, depth images, point clouds, and voxel grids more efficiently than for existing neural shape fields (see \Cref{fig:teaser}).
To extract point clouds or depth maps from \shortname, the forward map is a trivial recursive forward pass of the network.
To extract signed or unsigned distance fields, the forward map is a minimum operator over depth in all directions: $\min_{\forall i}[d(x, \hat{w}_i)]$.
To extract occupancy voxel grids, the forward map is a minimum operator over the intersection mask in all directions.
Finally, to extract meshes we propose an efficient variant of marching cubes that exploits the omnidirectional information and can handle open surfaces.


Experiments demonstrate that \shortname 
learns to overfit
high-quality shapes for single shape instances, 
with both open and closed surfaces.
\shortname also generalizes to a category of shapes when trained using an autodecoder framework~\cite{park2019deepsdf}.
Our method can also handle occlusion boundaries and surface frontiers.
We envision \shortname to be used in applications that require learning from and generating different 3D representations.
In summary, our contributions are:
\begin{packed_itemize}
\item A novel 3D representation, \textbf{Omnidirectional Distance Fields (ODFs)}, that returns the distance to the surface for any ray in 3D space. 
\item \textbf{\shortname}, a method for learning ODFs using a recursive neural network.
\item Forward mapping algorithms to efficiently extract common 3D representations like depth images, meshes, and point clouds from ODFs/NeuralODFs.
\end{packed_itemize}


%% file: content/02_relwork.tex
\section{Related Work}
\label{sec:rel}
In this brief survey of related work, we focus on learning-based methods (including neural shape fields~\cite{xie2021neural}) for 3D shape representations such as point clouds, voxels, meshes, parametric surfaces, level sets, multi-view images, and radiance fields.

\parahead{Point Clouds}
Point clouds are one of the most common 3D geometric data structures.
Therefore, representing and processing point clouds with neural networks has been an important topic of recent work~\cite{qi2017pointnet,wang2019dynamic,qi2017pointnet++,shoef2019pointwise}.
Early work used features extracted from permutation-invariant networks like PointNet together with an adversarial loss~\cite{achlioptas2018learning,li2018point}, while other methods use convolution-based architectures~\cite{gadelha2018multiresolution}.
Another approach is to start with a regular point set, sampled either on a grid~\cite{yang2018foldingnet} or from a normal distribution~\cite{yang2019pointflow}, and then advected them to appropriate 3D locations to represent the shape.
Point clouds can be transformed to other representations, but the process can be slow and inaccurate (\eg~Poisson surface reconstruction~\cite{kazhdan2006poisson}).
Different from these methods, \shortname can trivially produce point cloud outputs (\Cref{sec:forwardmaps}) that can be sampled at arbitrary resolutions in addition to producing other common representations.

\parahead{Voxels}
Voxels are another common 3D representation that are easy to store, query and manipulate.
Some methods extend the notion of 2D convolution to 3D voxels~\cite{choy20163d,liu2015deep,jimenez2016unsupervised}.
Using neural nets to estimate voxel occupancy is a common approach, for instance, in Occupancy Networks~\cite{mescheder2019occupancy}.
Since capturing small-scale detail is difficult with a global latent code, some methods divide the shape into smaller latent code grids~\cite{peng2020convolutional}.
Voxel grids generally lack the ability to capture small details at lower resolutions, while storage and compute requirements for higher resolutions grow exponentially.
They are also not conducive for transforming to other representations easily due to missing details.

\parahead{Meshes / Parametric Surfaces}
Meshes are perhaps the most common 3D data structure used widely in rendering and 3D reconstruction.
Since meshes are irregular, graph/geodesic convolution operations are a common choice within neural networks~\cite{bronstein2017geometric,yang2021continuous,guo20153d,yang2018foldingnet}.
Features learned on mesh vertices can directly be learned for use in downstream tasks~\cite{hanocka2019meshcnn,wiersma2020cnns,wang2018pixel2mesh,gkioxari2019mesh,dai2019scan2mesh}.
However, meshes have fixed topology and resolution, so some methods have focused on \emph{parametric surfaces} that can be sampled at any resolution~\cite{groueix2018papier,deprelle2019learning,lei2020pix2surf}.
While these methods are continuous, they have difficulty reconstructing shapes with varying topology.
Due to their popularity, well established methods exist for converting meshes to point clouds and depth images~\cite{perlin1989hypertexture,hart1996sphere}.

\parahead{Level Sets / Distance Fields}
Level sets such as signed distance fields (SDFs) are suitable for representing shapes of varying topology.
Multiple methods have demonstrated that SDFs can be learned with a neural network~\cite{park2019deepsdf,duan2020curriculum}, and can be further improved by regularizations (\eg~Eikonal property) specific to SDFs~\cite{gropp2020implicit}.
Marching Cubes~\cite{lorensen1987marching,liao2018deep} is a popular method for converting level sets to meshes.
However, SDFs are constrained to modeling watertight shapes limiting their use in open surfaces (\eg~garments).
Unsigned distance fields~\cite{chibane2020neural} (with Ball-Pivoting~\cite{bernardini1999ball}) and GIFS~\cite{ye2022gifs} are alternatives that support open surfaces and mesh generation.
To model high-quality shape, recent methods employ vector quantization together with transformers~\cite{yan2022shapeformer,mittal2022autosdf}.
Concurrent to our work, Probabilistic Directed Distance Fields~\cite{aumentado2021representing} were recently introduced and use a similar shape representation as us.
However, their focus is on differentiable rendering, and the use of regularization for handling discontinuities.
In contrast, we focus on efficient forward maps for shape reconstruction, and show that a recursive MLP is sufficient to learn ODFs.

\parahead{Multi-View Images / Radiance Fields}
While outside our scope, we note that ODFs are connected to networks that learn from multi-view images and are used to render photo-realistic novel views.
Methods like NeRF~\cite{mildenhall2020nerf}, SRNs~\cite{sitzmann2019scene}, VolSDF~\cite{yariv2021volume} and NeuS~\cite{wang2021neus} implicitly encode the scene radiance / geometry and require ray marching to extract depth maps.
Recently, Light Field Networks~\cite{sitzmann2021light} accelerated the neural rendering process by directly parameterizing the radiance observed by each ray.
Unlike Light Field Networks~\cite{sitzmann2021light} which focuses on scene radiance, our focus is on accurate 3D shapes.
Our method could be used to speed up rendering and efficiently extract depth maps.

%% file: content/03_method.tex
\section{ODF: Omnidirectional Distance Field}
\label{sec:ODF}
\begin{figure}
\begin{center}
\includegraphics[width=0.34\textwidth]{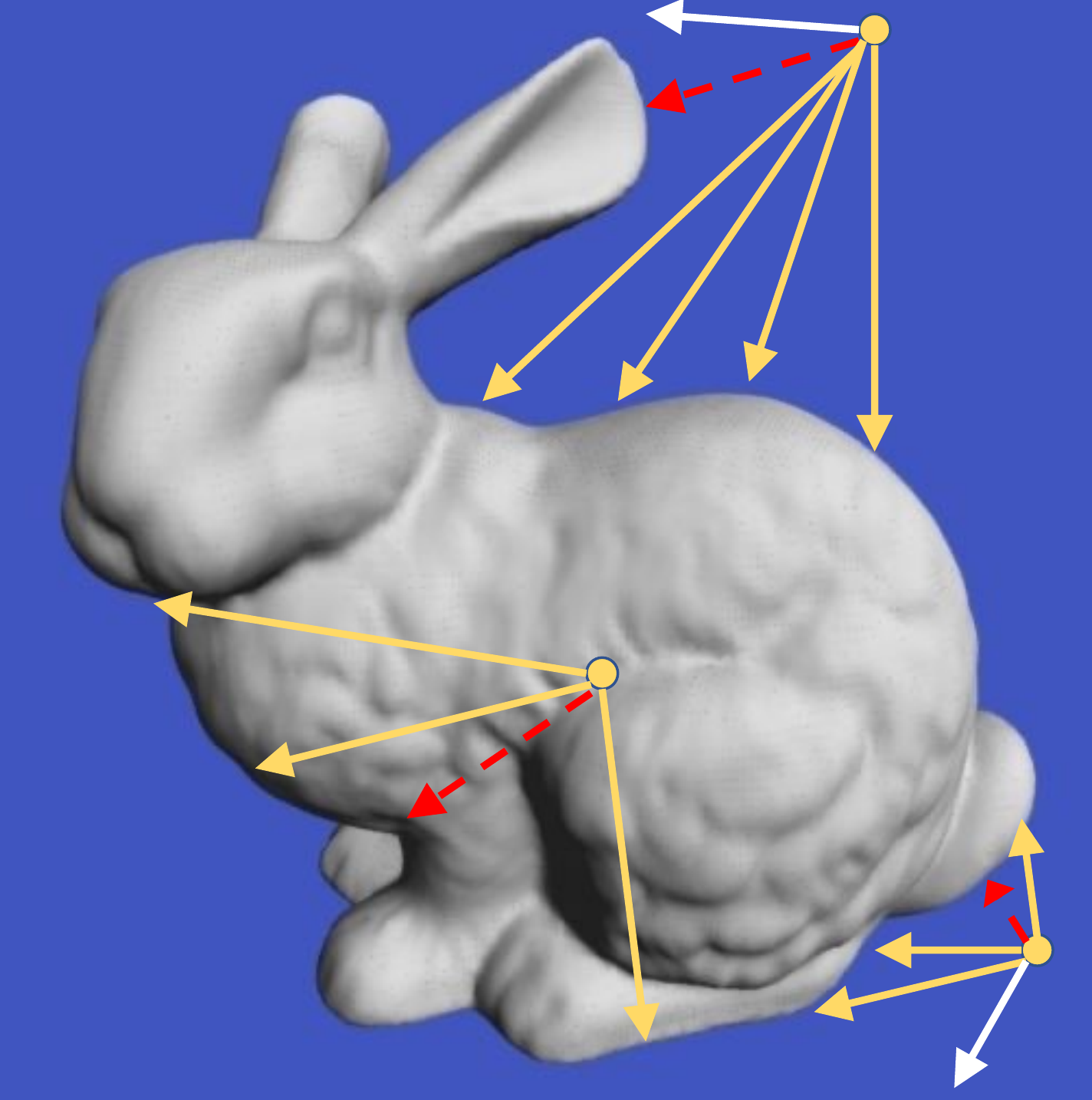}
\end{center}
\caption{An ODF maps all input rays $(\vec r = (x, \hat{w}))$ to the corresponding depths and intersections (white rays do not intersect, \colorbox{yellow!40}{yellow rays intersect}, \colorbox{red!40}{red rays model SDF/UDF}).}
\label{fig:odf}
\end{figure}

In this work, we propose a new 3D representation called Omnidirectional Distance Field (ODF) that is inspired by light fields~\cite{lumigraph,LightField} for modeling the 5-dimensional plenoptic function~\cite{bergen1991plenoptic,PlenopticFunction}.
Similar to a light field, an ODF is also a 5D function that implicitly encodes 3D geometry by storing the depth to a shape's surface along rays intersecting it.
Formally, we define an ODF as a function $\mathrm{ODF}(\vec r) = (d, p)$, which takes as input a vector defining the input ray $\vec r = (x, \hat{w})$, where $x$ denotes the ray starting point and $\hat{w}$ denotes a unit ray direction.
For each input ray, the ODF function outputs a scalar distance value ($d \in \mathbb{R^+}$) to the surface of the underlying geometry (see \Cref{fig:odf}).
The sign of the scalar distance specifies whether the sampled point is inside or outside the geometry (although our focus is primarily on \emph{unsigned ODFs}).
Since some rays will never intersect the shape, an ODF additionally stores a binary flag ($p \in \{0, 1\}$) to indicate intersection.
%
%
%
For ray $\vec r$ intersecting the surface (\ie~$p = 1$), the intersection point can be directly generated in $\mathbb{O}(1)$ complexity as: $x + \hat{w} d$.

ODFs store omnidirectional information at each 3D point and thus encode rich information about the
%
%
geometry. This enables us to build more efficient forward mapping functions compared to other shape representations.

\parahead{ODFs Generalize SDFs/UDFs}
Signed/unsigned distance field (SDF/UDF) at a given point $x$ refers to the shortest distance (along the surface normal) from the input point to the surface of the geometry.
The sign of the SDF specifies whether the sampled point is inside or outside the geometry, whereas a UDF is unsigned.
Compared to SDFs/UDFs, signed/unsigned ODFs model the distance to the surface along \textbf{any given input direction}, thus an \emph{SDF/UDF is a slice of the ODF along the surface normal direction}.

In this paper, we focus on \emph{unsigned} ODFs that encode a positive scalar distance for modeling open surfaces.
However, signed ODFs can also be constructed. \textcolor{black}{In Section \ref{sec:forwardmaps} we demonsrate that unsigned ODFs can still be used to approximate sign values on closed shapes.}

\parahead{Recursive Property}
\textcolor{black}{
SDFs are known to follow the Eikonal property, which states that the magnitude of the gradient of the signed distance field should satisfy $\| \nabla \mathrm{SDF}(x)\| = 1$.
Recent works \cite{icml2020_2086} leverage this property to regularize learned SDFs.
Since ODFs are also distance fields, they must satisfy a similar property which we refer to as the Recursive Property. For a fixed ray direction, the gradient of the ODF should be -1 as the position is moved in the ray direction and approaches the object surface. 
\[\nabla_{[\hat{w}, \vec{0}]}\mathrm{ODF}_{depth}(x, \hat{w})=-1\]
This property makes ODFs a recursive function of the depth value:
\begin{align}
    \mathrm{ODF}_{\mathrm{depth}}(x, \hat{w}) =  \mathrm{ODF}_{\mathrm{depth}}(x - \hat{w}, \hat{w})  + 1\nonumber
\end{align}
We use this recursive property to regularize our learned ODFs via data augmentation (see \Cref{sec:constructing}). Note that the Recursive Property does not hold across surface boundaries due to discontinuities in the ODF. This limitation does not affect our ability to use the Recursive Property for data augmentation.
}
\section{NeuralODF: Neural Omnidirectional Distance Fields}
\label{sec:neuralodf}
While an ODF can be stored and queried without using neural networks, this can be computationally expensive due to dimensionality and can result in low-quality shapes.
To capture details and to enable sampling continuously at arbitrary resolutions~\cite{scarselli1998universal}, we build upon recent successes in neural shape fields that learn shape representations implicitly~\cite{park2019deepsdf,groueix2018papier,mildenhall2020nerf}.
%
We first describe our procedure to learn \shortname, and then our forward mapping algorithms that use the \shortname to extract common 3D representations such as meshes, point clouds, and depth maps.

\begin{figure*}[t]
  \centering
  \includegraphics[width=\textwidth]{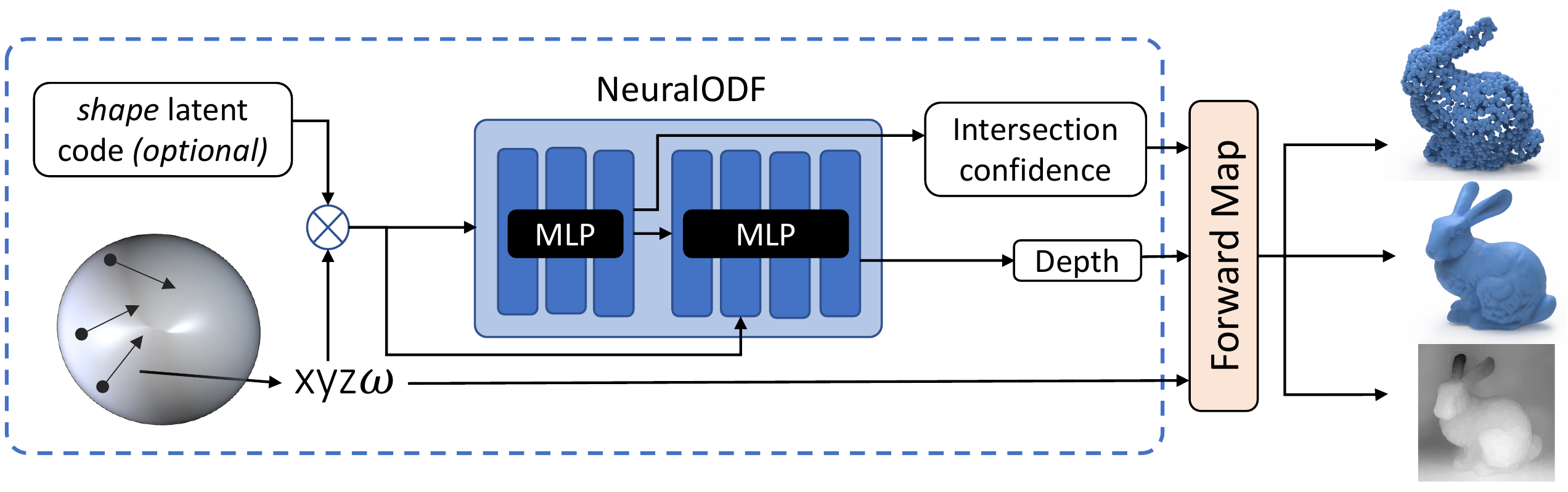}
  \caption{\textbf{Overview of NeuralODF:}
  The proposed neural omnidirectional distance field is trained
using a set of sampled 5D rays (directed arrows in the sphere above), and the associated distance and intersection labels as output.
For each sampled ray $\vec r = (x, \hat{w})$, the NeuralODF module predicts an intersection confidence and a depth-to-surface estimate.
We use a recursive inference procedure for higher quality shape reconstruction, where we recursively query the NeuralODF at multiple points along the input ray to predict the \textcolor{black}{surface distance and a binary flag indicating whether the ray intersects the object (see appendix for details)}. 
Once learned, we can map the learned ODF to 3D representations like meshes, point clouds, and depth maps.
\shortname can be used for both overfitting a single shape and generalizing to a category of shapes.
}
  \label{fig:architecture}
  \vspace{-5mm}
\end{figure*}

\subsection{Constructing \shortname}
\label{sec:constructing}
To learn ODFs with a neural network, we adopt a supervised training strategy.
Learning an ODF requires us to sample \emph{rays} around 3D shapes, build a neural network that can learn to estimate the depth and intersection of these rays, and propose a loss function to guide the training of the network.
Furthermore, we would like to support overfitting of single shapes, as well as generalization to categories of shapes.

\begin{figure*}[t]
  \centering
  \includegraphics[width=\textwidth]{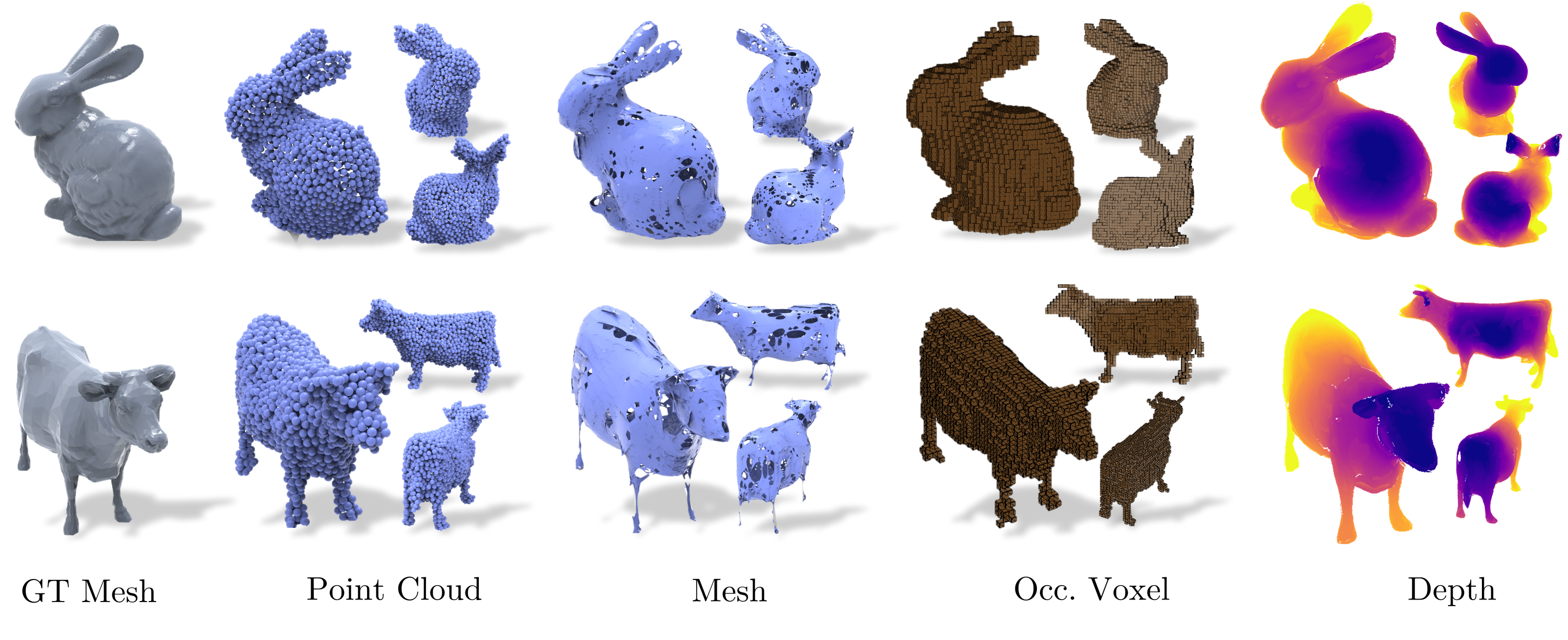}
  \caption{\textbf{Various 3D representations extracted from learned NeuralODF:} We showcase ODF's 3D fitting capability through a mesh $\longrightarrow$ ODF $\longrightarrow$ \{point cloud, mesh, voxel, depth\} transformation.}
  \label{fig:overfitting-exp}
  \vspace{-5mm}
\end{figure*}

\parahead{Sampling Rays for Learning}
%
%
As described in \Cref{sec:ODF}, rays are the fundamental building block of an ODF.
Thus, to learn an ODF, we first need to sample rays around shapes that we are interested in capturing.
The input to \shortname could be any one of the common shape representations like meshes, depth maps, or point clouds.
To generate training data from these input representations, 
we sample a set of 5D rays $(\vec r = (x, \hat{w}))$ with associated depth and intersection mask.
We  use well-established techniques to sample rays from common representations, for instance, fast rendering for meshes, pixel sampling for depth maps, or ray sampling around point clouds.
This allows us to obtain a balanced random sample of intersecting and non-intersecting rays distributed uniformly within an enclosed volume (sphere of radius $ = 1.3$).
We discuss the specific parameters in our data preparation procedure in \Cref{exp: data_preparation}.
Our focus is on unsigned ODFs for modeling open surfaces, so we only use positive scalars for depth.
Formally, each instance in our training data refers to a pair of ray starting point and unit direction: $(x, \hat{w})$, together with labels that store the depth to the surface ($d$), and intersection ($p$).
For rays not intersecting the 3D geometry, the depth value is set to $0.5$. \textcolor{black}{Since we clamp depth values to 0.5 when calculating the loss, this allows our network to predict any large depth value for non-intersecting rays without penalty.}
Our training rays are shown as directed arrows within a sphere in \Cref{fig:architecture}.


\parahead{Network Architecture}
Inspired by previous work that learn high dimensional functions~\cite{mildenhall2020nerf,park2019deepsdf}, \shortname uses an MLP architecture (see Figure~\ref{fig:architecture}) to learn ODFs.
It consists of eight feed-forward dense layers (with ReLU activation and Layernorm \cite{layernorm}).
The input to this network is a ray denoted by 3D position and a unit 3D direction vector (6D). \textcolor{black}{Though the input is 6D, the direction vector has only 2 degrees of freedom, so our input space corresponds to a 5D manifold.}
The output of this network is an intersection confidence value and a depth along the ray.
We output the intersection confidence after two layers in order to \emph{leave larger network capacity for the harder depth estimation task}.
We also include a skip connection of the input 6D coordinates (and latent code) for better depth estimation.

\parahead{Overfitting \& Generalization}
\shortname supports both overfitting a single shape or generalizing to a category of shapes.
For overfitting~(\Cref{sec:single_instance}), we directly train \shortname on a given shape's training data (rays).
To generalize to a category of shapes~(\Cref{sec:generalization}), we adopt the autodecoder framework~\cite{park2019deepsdf}.
During training, we jointly optimize the network weights and a 256-dimensional latent code for each instance in the shape category.
During inference, each latent code maps to a unique instance in the category and new latent codes can also be optimized for unseen instances.

\begin{figure*}[t]
  \centering
  \includegraphics[width=0.85\textwidth]{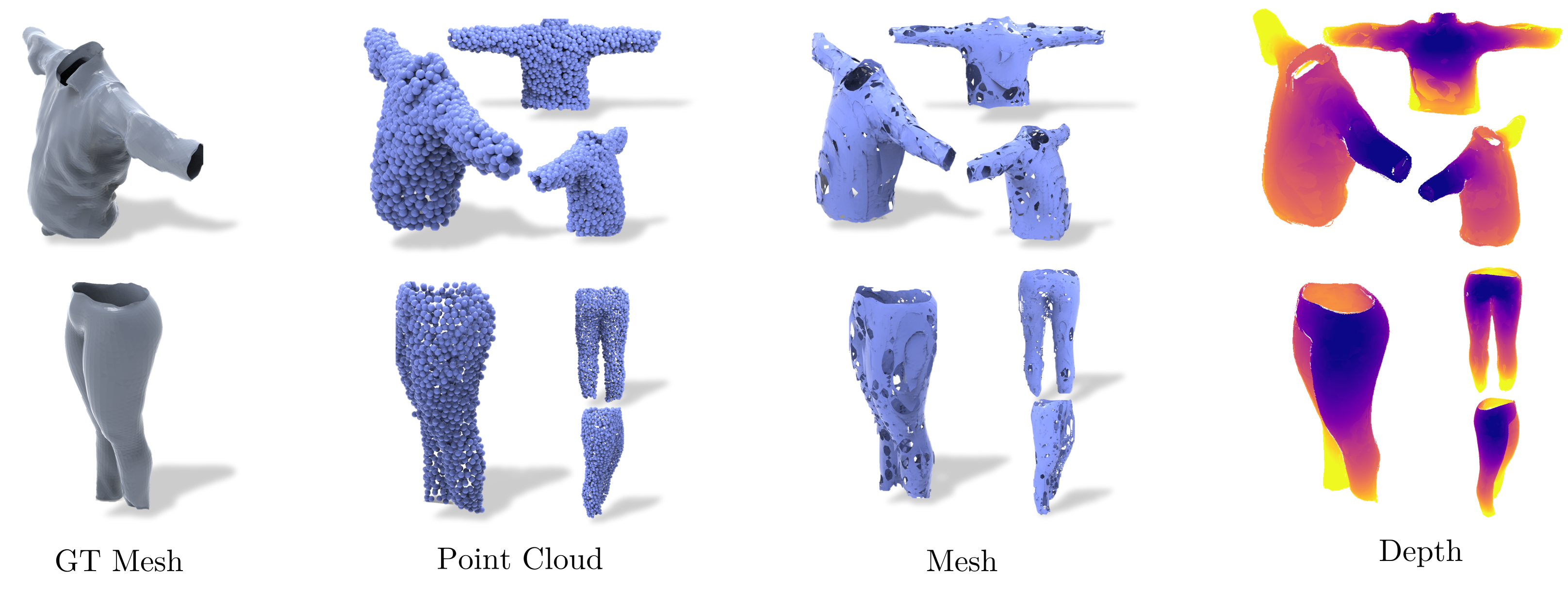}
  \caption{\textbf{Open surface extraction using NeuralODF:} Open-surface 3D garments reconstructed through a mesh $\longrightarrow$ ODF $\longrightarrow$ \{point cloud, mesh, voxel, rendered depth maps\} transformation.}
  \label{fig:opensurface}
  \vspace{-5mm}
\end{figure*}

\parahead{Recursive Inference}
During inference, a forward pass of \shortname can directly be used to obtain intersection and depth for any ray within the volume.
However, this can often result in noisy estimates, especially for rays originating far from the surface.
For higher shape quality, we introduce a recursive inference strategy that progressively refines the depth estimate along a given ray.
For an input ray $\vec r$, we first estimate $\mathrm{ODF}_{\mathrm{depth}}(x, \hat{w})$ and obtain a new 3D location $x_1$ that is the putative surface.
We then recurse inference from $x_1$ along the same direction $\hat{w}$ until the estimated depth difference is small (empirically set to 3 iterations, see supplementary).
This recursive procedure leads to higher quality shapes. Recursion is not used during training of the network.

\parahead{Loss Functions}
%
The training of ODF is guided by two loss terms: mean square error $(\ell_{\mathrm{depth}})$ with the GT for the depth prediction and binary cross-entropy loss $(\ell_{\mathrm{prob}})$ for the intersection confidence prediction task.
Like TSDFs~\cite{tsdf, park2019deepsdf}, we clamp both the predicted and ground truth depth to a max value of $0.5$, to better utilize the network capacity for near-surface points.
We follow~\cite{park2019deepsdf} to regularize the latent code with $\ell2$ norm for generalization $(\ell_{\mathrm{regularization}})$.
The overall loss is:
\begin{equation*}
\begin{aligned}
\label{eq:loss_all}
\ell_{all} = 
\lambda_1 \ell_{\mathrm{depth}} + \ell_{\mathrm{prob}} + \lambda_2*\ell_{\mathrm{regularization}}
\end{aligned}
\end{equation*}
We set $\lambda_1$ to $5$ and $\lambda_2$ to $0.0001$ in our experiments.


\subsection{Forward Mapping ODF to Common 3D Representations}
\label{sec:forwardmaps}

Defining geometry as an ODF allows us to extract various 3D representations, such as meshes, point clouds, depth maps and implicit functions at reduced time complexities and memory due to the omnidirectional information.
Once trained, we utilize the learned NeuralODF module to extract different shape representations at inference time as decribed below.

\parahead{ODF to Depth Maps}
%
Given an input camera viewpoint, we shoot multiple rays from the camera center in different directions corresponding to the underlying pixel locations and camera intrinsics.
A depth map can then be rendered by simply mapping the space of generated rays to the corresponding depth values obtained from the ODF. \textcolor{black}{The camera viewpoint is usually far from the center of the bounding space in order to include the whole object in the depth maps. Hence, we use 4 recursion steps for depth map rendering.}

\parahead{ODF to Point Clouds}
To extract a 3D point cloud from the ODF, we first sample 3D points within a sphere (radius=1.3 used in experiments) enclosing the underlying geometry.
These 3D points serve as the ray starting points.
\textcolor{black}{For each ray starting point, we then uniformly sample a ray direction and use our NeuralODF to estimate the depth with 3 recursion steps.}
Some of these rays do not intersect and are discarded.
Our resultant point cloud is simply the projected surface points corresponding to the intersecting rays. \textcolor{black}{Farthest point sampling can be applied as a post-processing step for more visually pleasing results.}

\parahead{ODF to Mesh}
The forward map for extracting a mesh out of a signed distance field is typically the marching cubes algorithm~\cite{lorensen1987marching}.
Marching cubes involve using the SDF values for each point in a lattice and then leveraging those values to determine the location of mesh faces (a mesh face occurs whenever there is a sign change in the SDF values of the neighbouring lattice points).
To generate a 3D mesh, the marching cubes algorithm performs a set of mesh intersection tests for each unit cube of the 3D lattice.
However, unlike SDF, the ODF value at a point on the lattice determines the depth to the geometric surface from that point.
This enables us to build a more efficient version of marching cubes that we call \textbf{Jumping Cubes}.
Based on the predicted depth estimates, Jumping Cubes skips a series of neighbouring cubes.
For instance, if depth to the surface along the x-direction from a lattice cube point is $k$, then it means that a mesh face would occur between the $\lfloor n+k \rfloor$ and $\lceil n+k \rceil$ lattice unit cubes and we can thus skip evaluating $k$ unit cubes between nth and $\lfloor n+k \rfloor$ cube.
Please see the supplementary document for details.

\begin{figure*}[t]
  \centering
  \includegraphics[width=1\textwidth]{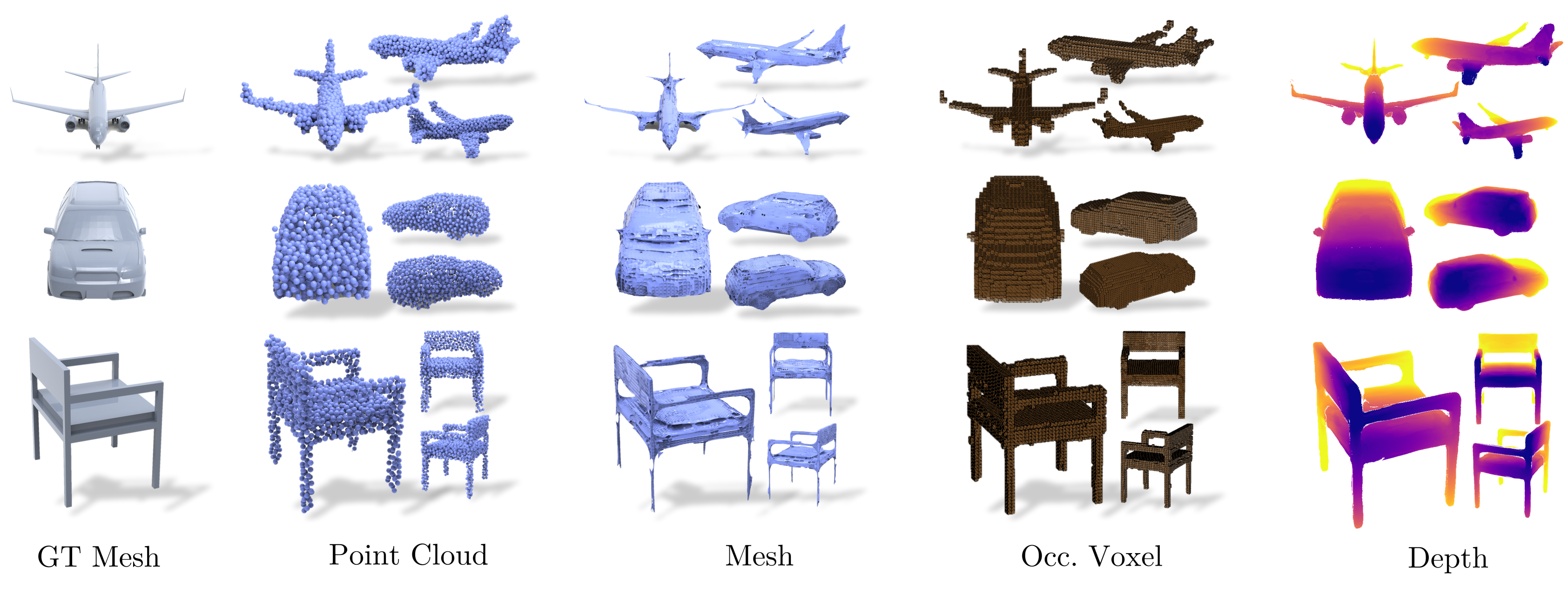}
  \vspace{-4mm}
  \caption{\textbf{Generalization of NeuralODF to Shapenet chairs and planes}}
  \vspace{-5mm}
  \label{fig:generalization}
\end{figure*}

\textcolor{black}{
\parahead{ODF to SDFs/UDFs}
ODFs can also be forward mapped to SDFs/UDFs since ODFs are a superset of these fields. Specifically, at a given position, $x$, the following equation relates UDFs to ODFs:
\[UDF(x) = \min_{\hat{w}} ODF_{depth}(x, \hat{w})\]
To compute the UDF at position $x$, we randomly sample $N$ different directions. We can then find the UDF value at that location by 
\[UDF(x) = \min_{\forall i}[ODF_{depth}(x, \hat{w}_i)].\] 
For SDFs we also need to compute the sign value. For points inside of a closed object, a ray going in any direction must encounter a surface, while this is usually not true of points outside of the object. Thus, if we query many ray directions from one viewpoint and they all intersect the surface, we take the point to be inside the object, otherwise, it is assumed to be outside the object. For a given point $x$, we use $N$ viewing directions to calculate the sign:
\[sign(x) = \begin{cases} -1 & \text{if } \min_{\forall i}[ODF_{intersect}(x, \hat{w}_i)] > 0.5\\ 1 & \text{otherwise}\end{cases}\]
We use this derivation of the SDF to produce occupancy voxel grids in Figure \ref{fig:overfitting-exp}.
}

\parahead{Time Complexity of ODF}
\textcolor{black}{With an ODF, we can retrieve a surface point with a single network query. Hence, the time complexities for rendering a point cloud of size $n$, a depth image with $h \times w$ pixels, or a voxel with grid of size $n \times n \times n$ are $\mathbb{O}(n)$, $\mathbb{O}(hw)$, and $\mathbb{O}(n^3)$, respectively. Furthermore, the Jumping Cubes Algorithm can be used to create a mesh from an $n \times n \times n$ lattice in $\mathbb{O}(n^2)$ time. Additional information about Jumping Cubes and its runtime can be found in the supplementary material.}

%% file: content/04_experiments.tex
\section{Experiment Details}
\label{sec:exp}

\subsection{Training Data Preparation} \label{exp: data_preparation}
To validate the efficiency of the learned ODF, we perform two sets of experiments: \textit{single-instance fitting experiments and generalization experiments.} In order to train the ODF for either of these experiments, we need to sample many ground-truth (GT) 5D rays, associated with the corresponding GT surface intersection probability and GT depth-to-surface estimates. While generating ground-truth SDF data requires a training instance to be a watertight 3D mesh \cite{park2019deepsdf}, the training instances for extracting ODF data (5D rays with depth and intersection estimates) need not be water-tight and can be in either mesh, point cloud or depth map representation. All our experiments are categorized based on the input and the output representations used (i.e. input representation $\longrightarrow$ ODF $\longrightarrow$ output representation). We now define the strategies to map different input representations to ODF:

\parahead{Multi-view depth maps to ODF} When training instances are available as a collection of 2.5D depth maps, the GT training ray data is simply the set of rays connecting the corresponding camera center to different depth-map pixel locations.  We used depth maps of resolution $256 \times 256$ for our experiments.
The camera centers for generating the depth maps are sampled on the surface of the enclosing sphere (radius $= 1.3$) and the camera's principle axis direction connects the camera center to the enclosing sphere center. 

\parahead{Mesh to ODF} When training instances are available as 3D meshes, we use two strategies to generate the training data: (a) For 60\% of the total sampled rays, we uniformly sample the ray starting point within the enclosing sphere and uniformly sample a ray direction from the surface of the unit sphere. (b) For 40\% of the rays, we first uniformly sample ray end points on the surface of the mesh and then sample ray start points inside the enclosing sphere. The ray direction is then simply the normalized vector connecting start and end points. For both the strategies, after sampling the ray start points and the directions, we determine the depth to surface (and hence recompute the end points) by performing ray-mesh intersection queries (using trimesh).


\parahead{Training data Augmentation} To further augment the training data, we leverage the recursive property of the ODF and perform the following augmentations: (a) Perturbation of the ray starting point along the chosen ray direction. (b) Perturbation of the ray end points along the chosen ray direction, and using them as the ray start points. The resulting ray direction in this case is opposite of the perturbation direction. The perturbations are restricted to be less than $0.1/0.01$ for overfitting/generalization. (c) Using the original ray end-points (surface points for intersecting rays) as the reference points. The new ray directions and start points are then sampled uniformly, centered at the reference points. The magnitude of the rays is restricted by $0.1/0.01$ for overfitting/generalization.

The last two augmentation strategies are used to sample 3D points near the surface of the input geometry. Please refer to supp. for ablation on these training data augmentation strategies.

\subsection{Datasets}

For the experiments involving fitting to a single 3D shape, we use the the commonly used CAD models in graphics to generate the input data: Stanford bunny, XYZ-RGB Asian dragon, cow and armadillo from Alec Jacobson's repository \cite{jacobsonCAD}. For demonstrating the efficiency of ODF in reconstructing open-surfaces, we also experiment on some garment instances \cite{bhatnagar2019multi}. 

For the generalization experiments, we train the proposed NeuralODF and the baselines on the Shapenet dataset \cite{chang2015shapenet} \textcolor{black}{and its watertight counterpart from DISN \cite{NIPS2019_8340} for the car, chair and airplane classes.} We leverage $1200$ CAD models as training instances and $100$ CAD models as test instances. Based on the input representation used for each experiment, we first map the CAD models to the corresponding input representation and then use the procedures mentioned in section \ref{exp: data_preparation} to map the input representation to the ODF training data.



\subsection{Metrics}

We evaluate both the intersecting probability and depth prediction of the generated ODFs. \textcolor{black}{If not mentioned otherwise, our GT point clouds are lifted from labeled ray data.}

\parahead{Metrics for evaluating depth prediction} 
We compute the chamfer distance (scaled by 1000) between the GT point cloud and the predicted point cloud (generated as the ray end-points of the intersecting rays). Finally, a f-score metric with threshold 0.005 is provided.

\parahead{Metrics for evaluating intersection probability prediction} For all input rays (intersecting and non-intersecting), we compare the GT and the predicted intersection probability through recall and f-score metrics. 

\section{Experimental Results}



\input{tables/single_inference}
\input{tables/shapenet_gen}

\subsection{Single-instance fitting experiments} \label{sec:single_instance} In this section, we test the capability of ODFs to fit to a single 3D training instance. We take ground-truth 3D shape of a single training object in an arbitrary input representation (depth map or mesh) and then map the input 3D representation to the ODF training data (section \ref{exp: data_preparation}). We then fit the NeuralODF module to the ODF data of the input 3D geometry. Once learned, we can then map the learned ODF to other 3D shape representations like meshes, point clouds, occupancy voxel grids and depth maps. In Figure~\ref{fig:overfitting-exp}, we perform the experiment of fitting a NODF to an input mesh, which is then forward mapped to various different output representations (mesh $\longrightarrow$ NODF $\longrightarrow$ \{mesh, point cloud, voxel, 2.5D depth map\}). From Figure ~\ref{fig:overfitting-exp}, we can see that the proposed NeuralODF module is able to reconstruct the \emph{finer  details (bunny ears, cow horns)} in the reconstructed shapes. 
In Figure~\ref{fig:opensurface}, we perform the same experiment for open-surface objects, which are difficult to model with signed-distance based approaches \cite{park2019deepsdf}. Thanks to the unsigned nature of the ODF, we are able to \emph{model the open-surfaces well and maintain the topology of the GT shape}. Table~\ref{table:sig_infer} quantitatively validates NeuralODF's performance. NeuralODF outperforms UDF within the same architecture, which confirms that ODF is a better representation.

\subsection{Generalization experiments} \label{sec:generalization}

NeuralODF is also able to generalize to unseen shapes. In this section, we train an autodecoder on 1200 ShapeNet objects \cite{chang2015shapenet} \textcolor{black}{(containing open surface shapes)} for each of the airplane, car, and chair classes. \textcolor{black}{Following NDF \cite{chibane2020neural}, we use ShapeNet processed by DISN \cite{NIPS2019_8340} as a closed surface shape counterpart for close surface baselines.} At inference time, we use 8 multi-view depth images from an unseen instance to optimize a new latent code. In Figure~\ref{fig:generalization} we use these latent codes to reconstruct unseen objects in various output representations. We can see that NeuralODF is able to capture important details like chair legs and jet engines on new instances. Classes with holes, like slatted chair backs, appear to be more difficult. Table~\ref{table:gen} assesses the performance of our autodecoder.

\subsection{Ablation study}

We perform several ablations to justify different choices made in the NeuralODF pipeline. 

\parahead{Ablation on recursive inference algorithm} Through Table~\ref{table:rec_infer} we ablate the recursive inference algorithm used to forward map ODFs to different 3D representations (section \ref{sec:forwardmaps}). From the table, we see that by \emph{using the recursive inference algorithm, we can reduce the chamfer error of the reconstructed point cloud by $12.4$\% and increase the F-score of the predicted intersection confidence by $22.6$\% (row 1 vs row 2)}. On further increasing the number of recursion steps (from 3 to 5) for ray marching using ODF, we notice an increase in the chamfer value. This is potentially because of the noise in the learned ODF and our recursive logic which could lead to the network predicting to the network predicting surfaces behind the viewpoint after many iterations.


\input{tables/iter_rec}
\input{tables/modality}

\paragraph{Ablation of input representations used for learning ODF:} 
As mentioned in Section \ref{exp: data_preparation}, ODF can be learned from various input representations (meshes, depth maps). To demonstrate this utility, we compare the single instance fitting capability using ODF training data generated from GT multi-view ($8$ in experiments) depth maps vs the ODF training data generated from GT meshes. From the table ~\ref{table:var_set}, we can see that the point clouds reconstructed using ODFs trained from meshes have significantly lower chamfer error than point clouds reconstructed using ODFs trained from depth maps. This is because input mesh representation provides more flexibility in sampling ray start points and directions (different for each sampled ray) compared to the ODF training data sampled from depth maps (ray start point same for all rays of one depth map). This can be fixed at the cost of sampling more multi-view depth maps.















%% file: tables/single_inference.tex
\begin{table}
\centering
\scalebox{0.88}{

\begin{tabular}{l|cc|cc}
\specialrule{.1em}{.1em}{.1em}
Method & \multicolumn{2}{c|}{Depth Metrics} & \multicolumn{2}{c}{Intersection Conf.} \\
& CD$\downarrow$  & F-score (\%) $\uparrow$ & Rec.(\%) & F-score (\%)  \\ 
\specialrule{.1em}{.1em}{.1em}
NeuralODF & \textbf{0.273}  & \textbf{27.02} & \textbf{96.97}  & \textbf{95.16}\\
UDF  & 0.857 & 15.80 & 86.94 & 80.77 \\ 
\specialrule{.1em}{.1em}{.1em}
\end{tabular}
}

\caption{Average results on single-object fitting. NeuralODF outperforms UDF with the same neural network architecture.}
\label{table:sig_infer}

\end{table}

%% file: tables/shapenet_gen.tex
\setlength{\tabcolsep}{4pt}
\begin{table}
\centering
\scalebox{0.92}{
\begin{tabular}{l|cc|cc}
\specialrule{.1em}{.1em}{.1em}
Category & \multicolumn{2}{c|}{Depth Metrics} & \multicolumn{2}{c}{Intersection Conf.} \\
& CD $\downarrow$  & F-score (\%) $\uparrow$  & Rec. (\%) $\uparrow$ & F-score (\%) $\uparrow$   \\
\specialrule{.1em}{.1em}{.1em}
Airplane & 1.123  & 12.33  & 97.42 & 85.92\\
Car & 1.284  & 13.54 & 98.43 & 94.56 \\ 
Chair & 1.246  & 12.34  & 97.08 & 90.29\\
\specialrule{.1em}{.1em}{.1em}
\end{tabular}
}
\caption{Generalization of NeuralODF on Shapenet. NeuralODF can generalize to unseen shapes with updated latent code.}
\label{table:gen}

\end{table}

%% file: tables/iter_rec.tex
\setlength{\tabcolsep}{4pt}
\begin{table}
\centering
\scalebox{1.1}{
\begin{tabular}{c|c|cc}
\specialrule{.1em}{.1em}{.1em}
Iter. & \multicolumn{1}{c}{Depth Metrics} & \multicolumn{2}{|c}{Intersection Conf. Metrics} \\
& CD $\downarrow$  & Rec. $\uparrow$ (\%) & F-score (\%) $\uparrow$  \\ 
\specialrule{.1em}{.1em}{.1em}
1  & 0.397  & 59.23  & 72.59\\
3  & 0.273  & 96.97 & 95.16\\
5  &  0.313 & 97.21 & 95.19\\
\specialrule{.1em}{.1em}{.1em}
\end{tabular}
}
\caption{Iterations of Recursive inference. We can reduce the chamfer distance of
the reconstructed point cloud and increase the
F-score throguh recursive inference. An iteration between 3 and 5 is a good choice.}
\label{table:rec_infer}
\end{table}

%% file: tables/modality.tex
\setlength{\tabcolsep}{4pt}
\begin{table}
\centering
\scalebox{0.95}{
\begin{tabular}{c|c|cc}
\specialrule{.1em}{.1em}{.1em}
Training Set & \multicolumn{1}{c}{Depth Metrics} & \multicolumn{2}{|c}{Intersection Conf. Metrics} \\
& CD $\downarrow$  & Rec. (\%) $\uparrow$ & F-score (\%) $\uparrow$  \\ \specialrule{.1em}{.1em}{.1em}
Mesh          & 0.273  & 96.97 & 95.16 \\ 
Depth Images  & 1.557  & 93.58 & 89.80\\
\specialrule{.1em}{.1em}{.1em}
\end{tabular}
}
\caption{NeuralODF with various fidelity. ODF can be learned
from various input representations.}
\label{table:var_set}
\end{table}

%% file: content/05_endmatter.tex
\section{Conclusion}
\label{sec:conclusion}
In this paper, we introduced a new 3D shape representation that we call Omnidirectional Distance Fields (ODFs).
ODFs represent shape by encoding the distance and visibility of rays originating at any 3D location and pointed in any direction.
Because of the omnidirectional information, ODFs make it more efficient to transform between different representations using forward maps.
Since ODFs are high dimensional and expensive, we use a neural network, \shortname to learn the ODF implicitly.
We show that an MLP architecture can effectively learn an ODF and can be used to query the ODF continuously at arbitrary resolutions.
Results show that we can capture high-quality shapes with \shortname and can efficiently forward map to common 3D representations like point clouds, depth images, meshes and voxel grids.

\parahead{Limitations}
Our method has several limitations which we plan to address in future work.
First, ODFs are high dimensional fields that are harder to learn than standard SDFs or UDFs, yet we show that a standard MLP can learn it with a recursive strategy.
Our Jumping Cubes algorithm works reliably for close shapes but can sometimes have trouble with open surfaces since it does not know which holes are spurious.
\shortname does not beat other methods in shape representation quality, but our goal is present it as an alternative option.

\parahead{Societal Impacts}
AI systems that use \shortname for shape representation can suffer from bias introduced by the dataset we use.
More diverse shapes will have to be introduced to avoid such biases.
Training \shortname uses several GPU-days of compute that results in energy consumption.
Future work should mitigate the large energy usage.

%% file: content/99_appendix.tex
\section{Abstract}
In this supplementary material, we first showcase the training procedure and results when the NeuralODF was trained with depth images  as the input representation (depth images $\longrightarrow$ ODF $\longrightarrow$ point cloud). Next, we describe in detail the core Jumping Cubes and the recursive marching algorithms. Finally, we showcase additional generalization results of NeuralODF on the Shapenet categories.

\section{Training with depth images}
\label{data_form}

\begin{figure}[thb]
  \centering
  \includegraphics[width=\linewidth]{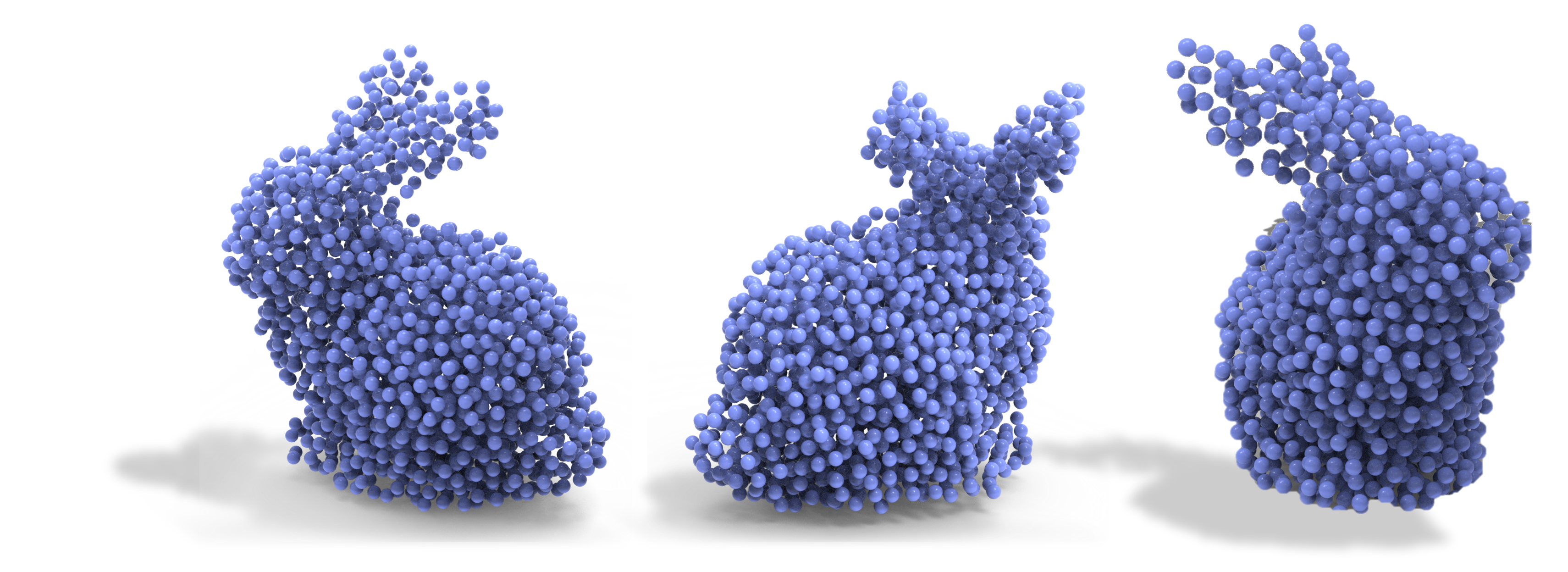}
  \caption{Point cloud reconstructions of the Stanford Bunny, retrieved from a NeuralODF trained on 32 depth images.}
  \label{fig:depth-image-training}
\end{figure}

The results shown for our overfitting and generalization experiments with NeuralODF were trained using mesh data. However, since the basic input of NeuralODF is a ray, we are also able to train on multi-view depth images and point clouds with normals. The details for how training data is generated from these representations is in the Training Data Preparation section of the paper. While point clouds with normals allow us to sample rays in a manner similar to meshes, depth images give us a ray bundle that provides a more biased sampling of our input domain. In Figure~\ref{fig:depth-image-training} we show a pointcloud reconstruction obtained from a NeuralODF trained on 32 depth images from cameras sampled uniformly from a sphere around the Stanford Bunny. These results show that we are still able to accurately learn an ODF when using depth images instead of a more complete representation such as a point cloud or a mesh.

\section{Jumping Cubes}
\label{jump_cube}

\subsection{Mesh extraction on open surfaces}
The Marching Cubes algorithm is commonly used to extract a mesh surface from implicit functions. It achieves this by evaluating the implicit function at the vertices of a 3D lattice grid. Each cube within the lattice can then be triangulated according to the sign of the function at each corner of the cube. As NeuralODF is an unsigned representation, this technique is not possible for us. However, by predicting surface depth, we can estimate which cube edge will intersect the object surface. An edge is determined to be intersected when the predicted depth at the vertex at one end of the edge is less than the length of the edge itself. Then, we can triangulate each cube based on which of its 12 edges are intersected by the object surface. The Marching Cubes algorithm has $2^8$ cases which can be reduced down to 14 base cases by removing configurations that are equivalent up to some rotation or change of sign. Similarly, for Jumping Cubes, we find that there are $2^{12}$ cases which can be reduced down to 218 cases when we account for rotations. In order to find the correct triangulation for each cube and produce a mesh, Jumping Cubes requires a binary array that indicates whether the surface intersects each cube edge on a 3 dimensional lattice grid. After extracting the surface with Jumping Cubes we run Laplacian smoothing. 

\subsection{Efficient jumping cubes with NeuralODF}

By predicting surface depth, ODFs allow us to efficiently compute the intersection array necessary to apply Jumping Cubes. In an $n \times n \times n$ lattice grid, there are $3n^3$ cube edges, or $n^3$ edges for each axis.  On one side of the lattice grid, there are $n^2$ points, each positioned at the head of a column of $n$ cube edges. For each of these columns, we use Algorithm ~\ref{alg:jc} to efficiently compute the depth value for each vertex in the column. The number of ODF queries this algorithm makes is dependent on the number of object surfaces the column passes through, not on the resolution of the grid. Therefore, we can run Jumping Cubes with only $\mathbb{O}(n^2)$ network queries, whereas Marching Cubes requires $\mathbb{O}(n^3)$ network queries.\\

\begin{algorithm}[hbt!]
\caption{ODF single-column inference for jumping cubes}
\label{alg:jc}
\begin{algorithmic}[1]
\State $x \gets$ Starting position
\State $\hat{w} \gets$ Viewing direction
\State $b \gets$ Steps adjustment
\State $s \gets$ Distance between vertices in the grid
\\
\State AllDepths = []
\While{Not at end of column}
    \State depth, intersection = $f(x, \hat{w})$
    \If{no intersection}
    \State set depth to a default value (0.5)
    \EndIf
    \State steps $\gets \max (1, \lfloor \frac{\text{depth}}{s} - b\rfloor)$
    \State AllDepths += [depth + $\hat{w}\cdot s\cdot 0$), ..., depth + $\hat{w}\cdot s \cdot$(steps-1)]
    \State $x \gets x + \hat{w}\cdot s \cdot\text{steps}$ 
\EndWhile
\Return AllDepths
\end{algorithmic}
\end{algorithm}

\begin{figure}
  \centering
  \includegraphics[width=\linewidth]{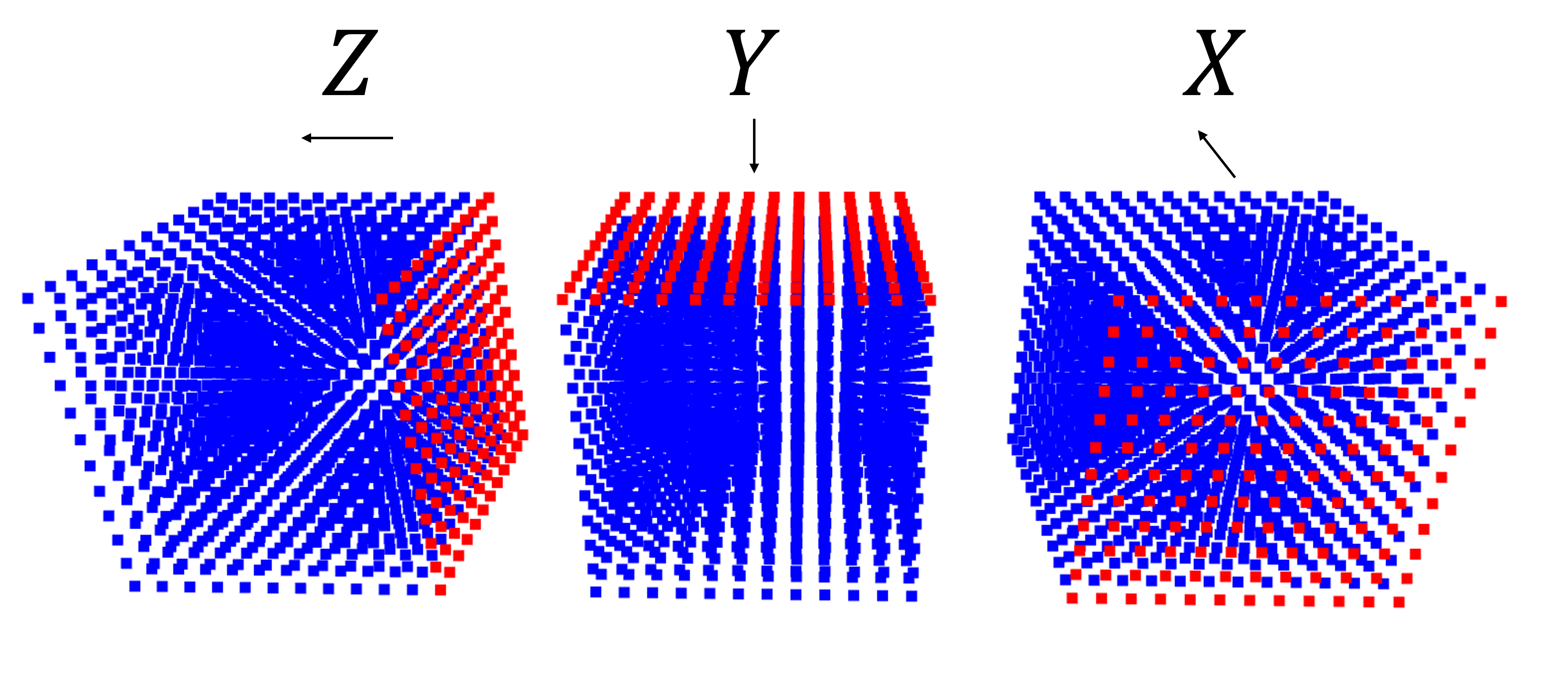}
  \caption{For Jumping Cubes, we must evaluate the depth at each lattice point, for each of the 3 axis-aligned viewing directions. Instead of directly querying each point, we start at the red points and jump down the column of points according to the predicted depth.}
  \label{fig:jc-lattice}
\end{figure}

\section{Recursive inference}
\label{rec_inf}
In this section, we discuss the details of the recursive inference. Firstly, given a starting position $x$ and viewing direction $\hat{w}$, we clamp the predicted depth $d$ since the depth larger than the clamping value $\psi$ is not constrained by the loss function (see algorithm~\ref{alg:infer} and~\ref{alg:infers} line 1-7). Secondly, we run the inference $n-1$ times. For each iteration, we check the forward and backward direction inference except for the first iteration and pick the minimum one to prevent overshooting the point $x$ (see algorithm~\ref{alg:infers} line 8-17). Finally, an additional forward and backward direction inference is used to generate the mask from the depth branch (see algorithm~\ref{alg:infers} line 18-22). 

\begin{algorithm}[hbt!]
\caption{An inference}
\label{alg:infer}
\begin{algorithmic}[1]
\State $x \gets$ Starting position
\State $\hat{w} \gets$ Viewing direction
\State $\psi \gets$ Clamping value
\State $d, m = f(x, \hat{w})$\\
\Return $min(d, \psi)$, $m$
\end{algorithmic}
\end{algorithm}

\begin{algorithm}[hbt!]
\caption{Recursive Inference for Neural ODF}
\label{alg:infers}
\begin{algorithmic}[1]
\State $x \gets$ Starting position
\State $\hat{w} \gets$ Viewing direction
\State $n \gets$ Number of recursive calls
\State $\tau \gets$ Surface margin
\\
Get forward direction inference by Algorithm 2
\State $x \gets x + d_{\text{forward}} \cdot \hat{w}$
\State $m \gets m_{\text{forward}}$
\State $n \gets n-1$
\While{$n \neq 0$}\\
Get forward and backward direction inference by Algorithm 2
\If{$d_{\text{forward}} \leq d_{\text{backward}}$}
     \State $x \gets x + d_{\text{forward}} \cdot \hat{w}$
     \State $m \gets m_{\text{forward}}$
\Else
     \State $x \gets x + d_{\text{backward}} \cdot -\hat{w}$
     \State $m \gets m_{\text{backward}}$
\EndIf
\State $n \gets n - 1$
\EndWhile \\
Get forward and backward direction inference by Algorithm 2
\If{$d_{\text{forward}} \leq d_{\text{backward}}$}
     \State $m \gets m\cap d_{\text{forward}}<\tau$  
\Else
     \State $m \gets m\cap d_{\text{backward}}<\tau$ 
\EndIf
\Return $x$, $m$
\end{algorithmic}
\end{algorithm}





\section{Ablation study of data augmentation}
\label{aug_ablation}

The recursive property of ODFs enables efficient data usage through dataset augmentation. Table ~\ref{table:aug} shows that learning an ODF without using any data augmentations is a challenging task. However, utilizing the recursive property efficiently through dataset augmentation can greatly improve the quality of our NeuralODF. The Table ~\ref{table:aug} results also show that using Aug-a alone results in significantly better quality than using Aug-b or Aug-c alone. We hypothesize that this may arise from the fact that Aug-b and Aug-c may include some error in the depth labels when non-target surfaces appear within the augmentation margin. Nevertheless, enabling all three augmentation strategies still yields the best reconstruction and improves the Chamfer Distance by 0.204 over the Aug-a only model.
\input{tables/aug}

\section{Additional Results}
In our generalization experiments, we used an autodecoder framework to train NeuralODF to represent object classes. Here we show reconstructions of additional instances from from these classes. Each instance shown here was not present in the training data, but was fit to the trained model by optimizing a latent code.

\begin{figure*}[h]
  \centering
  \includegraphics[width=0.8\linewidth]{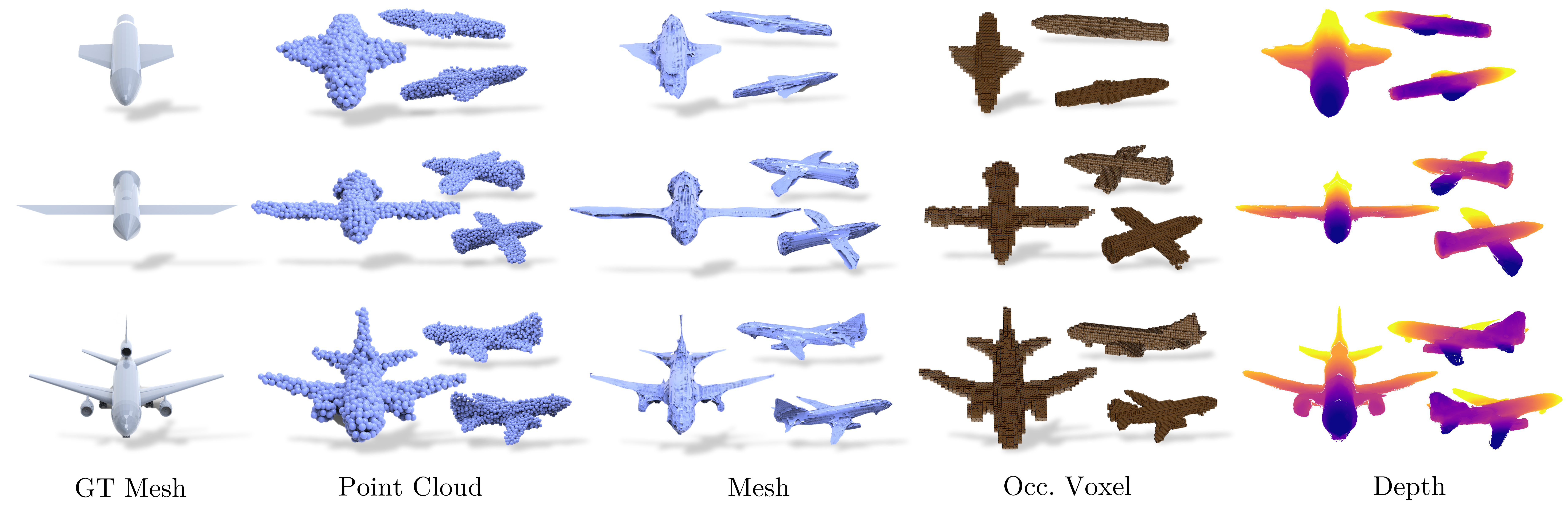}
  \caption{Reconstruction of unseen instances of the Shapenet airplanes class.}
\end{figure*}

\begin{figure*}[h]
  \centering
  \includegraphics[width=0.8\linewidth]{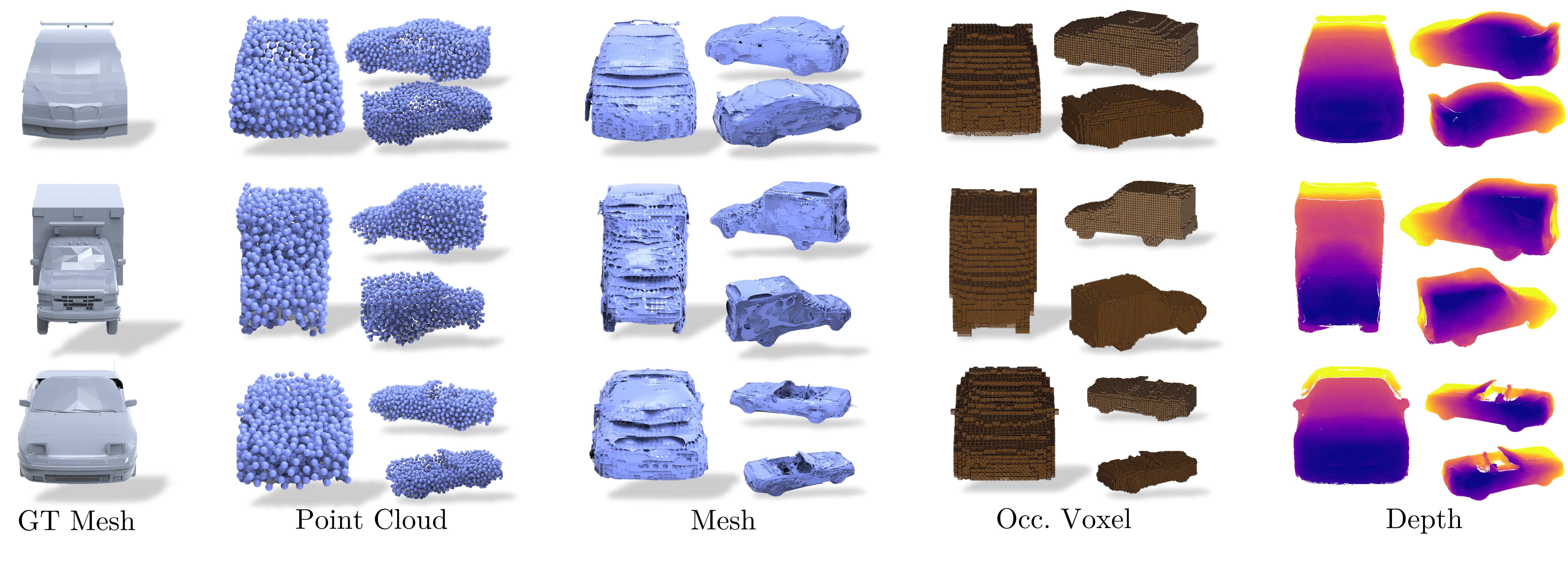}
  \caption{Reconstruction of unseen instances of the Shapenet cars class.}
\end{figure*}

\begin{figure*}[th]
  \centering
  \includegraphics[width=0.8\linewidth]{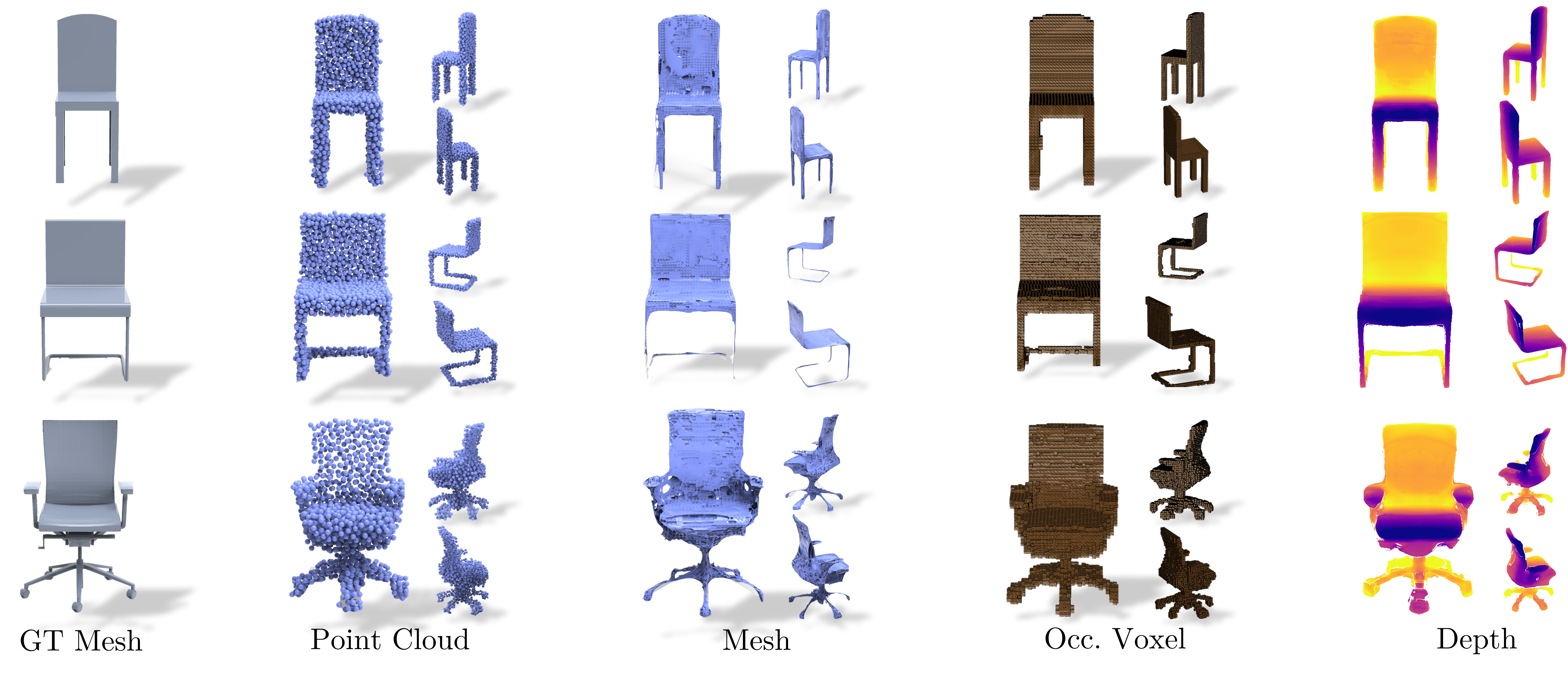}
  \caption{Reconstruction of unseen instances of the Shapenet chairs class.}
\end{figure*}

%% file: tables/aug.tex
\begin{table}[h!]
\centering
\scalebox{0.85}{
\begin{tabular}{ccc|c|cc}
\hline
Aug-a & Aug-b & Aug-c & \multicolumn{1}{c}{Depth Metrics} & \multicolumn{2}{|c}{Intersection Conf. Metrics} \\
& & & CD $\downarrow$  & Rec. (\%) $\uparrow$ & F-score (\%) $\uparrow$ \\
\specialrule{.1em}{.1em}{.1em}
& & & 5.5614  & 44.05& 58.62\\
$\checkmark$ & & & 0.4775   & 90.60& 92.87\\
& $\checkmark$ & & 0.8383  & 87.29& 84.95\\
& & $\checkmark$ & 0.8460   & 89.61& 85.34\\
$\checkmark$ & $\checkmark$ & $\checkmark$ & 0.2731  & 96.97& 95.16\\
\hline
\end{tabular}
}
\caption{Ablation Study for dataset augmentation. See Section 5.1 for the details.}
\label{table:aug}
\end{table}